\documentclass[review]{elsarticle}
\biboptions{authoryear}

\usepackage[T1]{fontenc}
\usepackage{fix-cm}
\usepackage{placeins,float}
\usepackage{ragged2e}
\usepackage{booktabs,siunitx}
\usepackage{tabularx,threeparttable}
\usepackage[font=small,labelfont=bf,labelsep=period]{caption}
\usepackage{makecell}
\usepackage{amsmath,amssymb,bm}
\usepackage{newtxtext,newtxmath}
\usepackage{multirow}
\usepackage{graphicx,verbatim}
\usepackage{pdfpages}

\usepackage{hyperref}
\hypersetup{unicode=true}

\journal{Human Brain Mapping}

\begin{document}
\begin{frontmatter}

\title{Improved Multiscale Structural Mapping with Supervertex Vision Transformer for the Detection of Alzheimer's Disease Neurodegeneration}

\author[aff1]{Geonwoo Baek}
\author[aff2,aff3,aff4]{David H. Salat}
\author[aff1]{Ikbeom Jang\corref{cor1}}
\author{for the Alzheimer’s Disease Neuroimaging Initiative}

\address[aff1]{Department of Computer Science \& Engineering, Hankuk University of Foreign Studies, Seoul, Republic of Korea}
\address[aff2]{Athinoula A. Martinos Center for Biomedical Imaging, Department of Radiology, Massachusetts General Hospital, Charlestown, MA, USA}
\address[aff3]{Department of Radiology, Harvard Medical School, Boston, MA, USA}
\address[aff4]{Neuroimaging Research for Veterans (NeRVe) Center, VA Boston Healthcare System, Boston, MA, USA}

\cortext[cor1]{Corresponding author}
\ead{ijang@hufs.ac.kr}

\begin{abstract} % 217 words - 250 word limit
Alzheimer’s disease (AD) confirmation often relies on positron emission tomography (PET) or cerebrospinal fluid (CSF) analysis, which are costly and invasive. Consequently, structural MRI biomarkers such as cortical thickness (CT) are widely used for non-invasive AD screening. Multiscale structural mapping (MSSM) was recently proposed to integrate gray–white matter contrasts (GWCs) with CT from a single T1-weighted MRI (T1w) scan. Building on this framework, we propose MSSM+, together with surface supervertex mapping (SSVM) and a Supervertex Vision Transformer (SV‑ViT). 3D T1w images from individuals with AD and cognitively normal (CN) controls were analyzed. MSSM+ extends MSSM by incorporating sulcal depth and cortical curvature at the vertex level. SSVM partitions the cortical surface into supervertices (surface patches) that effectively represent inter- and intra-regional spatial relationships. SV‑ViT is a Vision Transformer architecture operating on these supervertices, enabling anatomically informed learning from surface mesh representations. Compared with MSSM, MSSM+ identified more spatially extensive and statistically significant group differences between AD and CN. In AD vs. CN classification, MSSM+ achieved a 3\%p higher area under the precision–recall curve than MSSM. Vendor-specific analyses further demonstrated reduced signal variability and consistently improved classification performance across MR manufacturers relative to CT, GWCs, and MSSM. These findings suggest that MSSM+ combined with SV‑ViT is a promising MRI-based imaging marker for AD detection prior to CSF/PET confirmation.
\end{abstract}

\begin{keyword}
Multiscale Structural Mapping \sep Gray-to-white matter contrast \sep Alzheimer's disease \sep Supervertices \sep Imaging biomarker \sep MRI
\end{keyword}

\end{frontmatter}

\section{Introduction}

Alzheimer’s disease (AD) is the most common cause of dementia worldwide, with neuropathological processes initiating years before the onset of clinical symptoms \citep{jack2010a,jack2010b,bateman2012}. Detection is therefore critical for enabling preventive interventions and reducing disease burden. The National Institute on Aging and Alzheimer’s Association (NIA--AA) research framework defines disease status along the amyloid/tau/neurodegeneration (A/T/N) scheme, in which structural magnetic resonance imaging (MRI) serves as a non-invasive indicator of N \citep{frisoni2010,jack2018}. Accordingly, structural MRI biomarkers such as hippocampal volume and cortical thickness (CT) \citep{tosun2021} are used for screening prior to CSF/PET confirmation in research and clinical workflows. Plasma blood-based biomarkers (BBMs), such as the phosphorylated-tau 217/amyloid-$\beta$42 ratio, have recently emerged. However, they lack anatomical specificity and have been validated primarily in adults 55 years and older, with significantly reduced accuracy in individuals aged 80 years and older \citep{ashton2024,hansson2022,janelidze2021,cullen2023,palmqvist2025,hazan2025}.

To improve sensitivity, multiscale structural mapping (MSSM) \citep{jang2022} was introduced to integrate CT with gray-to-white matter contrasts (GWCs), and to project these high-dimensional vertex-wise features into a lower-dimensional latent space using partial least squares-discriminant analysis (PLS-DA) \citep{lee2018}. GWCs reflect microstructural boundary integrity and tissue contrast at the gray–white interface \citep{salat2009,salat2011,westlye2010} and were implemented in MSSM as multiple depth-wise intensity contrasts, which are sensitive to age- and disease-related changes. Consistent with this view, a recent study reported that reduced GWCs are associated with markers of neurodegeneration, amyloid burden, and cognitive impairment across independent cohorts, supporting GWCs as a sensitive indicator of cortical tissue integrity \citep{xu2024}. As a proof-of-concept, MSSM was initially evaluated on Siemens data, underscoring the need to establish cross-vendor generalizability. Recent work further extended MSSM to derive structural indicators of $\beta$-amyloid neuropathology in preclinical AD, showing that MSSM signals are regionally altered and closely aligned with amyloid PET burden even when cortical thickness or hippocampal volume alone are not \citep{jang2024}.

Beyond CT and GWCs, additional vertex-wise morphometric features may capture complementary aspects of cortical architecture relevant to AD. Sulcal depth (Sulc) indexes cortical folding and sulcal widening associated with tissue loss \citep{schaer2008,thompson2004}, whereas cortical curvature (Curv) quantifies local shape complexity and has been linked to cortical thinning and gyral collapse in neurodegeneration \citep{hogstrom2013}. The white-matter Jacobian determinant (Jacob) reflects localized expansion or contraction and has been used extensively in deformation-based morphometry \citep{ashburner2000,gaser2001}. The local gyrification index (LGI) quantifies folding extent and has been associated with developmental and pathological variation in cortical complexity \citep{schaer2008}. Incorporating such intensity and geometry measures into a unified surface representation may enhance sensitivity to subtle cortical alterations. At the same time, it remains pivotal to identify compact feature subsets that maximize incremental value, as not all morphometric descriptors contribute equally in multivariate classification.

In parallel with advances in feature design, deep learning methods have increasingly been applied to surface-based neuroimaging data. Graph convolutional approaches such as Hypergraph Neural Network+ (HGNN+) \citep{feng2022}, Spline-based Graph CNN (SplineCNN) \citep{fey2018}, SpiralNet++ \citep{gong2019}, and Surface Vision Transformer (SiT) \citep{dahan2022} have been applied to cortical meshes.

We developed surface supervertex mapping (SSVM) to extend this line of work by mapping cortical surfaces into non-overlapping supervertices (SVs). Because cortical measurements reside on a non-Euclidean two-dimensional manifold embedded in three dimensions, projecting them onto planar grids distorts surface areas and neighborhood relations. Standard 2D encoders, which assume a Euclidean lattice, often require ad-hoc unwrapping or resampling that can blur vertex-wise signals. We therefore designed the Supervertex Vision Transformer (SV-ViT) to directly use SVs as inputs to enable region-of-interest (ROI)-informed learning.

In this study, we extended MSSM by systematically evaluating vertex-wise cortical features (GWCs, CT, Sulc, Curv, Jacob, and LGI) across multi-vendor T1-weighted datasets. We identified a compact subset, referred to as MSSM+, comprising GWCs, CT, Sulc, and Curv. We further proposed SSVM and designed SV-ViT, benchmarking them against the latest models for AD versus cognitively normal (CN) classification. Together, these advances provide evidence for generalizability across Siemens, GE, and Philips scanners and support MSSM+ with SSVM and SV-ViT as a promising surface-based framework for MRI-based detection of AD prior to CSF/PET confirmation in AD neurodegeneration.

\section{Material and methods}

% GE 파라미터 2종류 슬래시로 표기

\begin{table}[H]
  \captionsetup{position=top, justification=justified, singlelinecheck=false}
  \caption{T1w acquisition parameters by dataset, vendor, and protocol (3T, sagittal, 3D). ADNI scans were accelerated, and OASIS scans were fully sampled. Values summarize representative acquisition parameters of the retained scans in the final analyzed dataset; ranges denote within-phase variability.}
  \label{tab:mri_params}
  \setlength{\tabcolsep}{5pt}
  \renewcommand{\arraystretch}{1.12}
  \small
  \begin{tabularx}{\textwidth}{@{\hspace{2pt}}
    c
    >{\centering\arraybackslash}p{1.3cm}
    >{\centering\arraybackslash}p{1.6cm}
    >{\centering\arraybackslash}p{1.1cm}
    >{\centering\arraybackslash}p{1.1cm}
    >{\centering\arraybackslash}p{1.1cm}
    >{\centering\arraybackslash}p{2.25cm}}
    \toprule
    Dataset & Vendor & Protocol & TR (ms) & TE (ms) & TI (ms) & Voxel size (mm$^3$) \\
    \midrule
    ADNI--GO & Siemens & MPRAGE          & 2300    & 3.0       & 900 & $1.0\times1.0\times1.2$ \\
             & Philips & MPRAGE          & 7       & 3.1       & 900 & $1.0\times1.0\times1.2$ \\
    ADNI--2  & Siemens & MPRAGE          & 2300    & 3.0       & 900 & $1.0\times1.0\times1.2$ \\
             & GE      & IR-FSPGR        & 7--8    & 2.8--3.2  & 400 & $1.0\times1.0\times1.2$ \\
             & Philips & MPRAGE          & 7       & 3.1       & 900 & $1.0\times1.0\times1.2$ \\
    ADNI--3  & Siemens & MPRAGE          & 2300    & 3.0       & 900 & $1.0\times1.0\times1.0$ \\
             & GE      & IR-FSPGR         & 7--8    & 2.9--3.2  & 400 & $1.0\times1.0\times1.0$ \\
             & Philips & MPRAGE          & 6--7    & 2.9--3.0  & 900 & $1.0\times1.0\times1.0$ \\
    ADNI--4  & Siemens & MPRAGE          & 2300    & 2.9--3.2  & 900 & $1.0\times1.0\times1.0$ \\
             & GE      & IR-FSPGR        & 7       & 2.9--3.1  & 400 & $1.0\times1.0\times1.0$ \\
             & GE      & MPRAGE          & 2300    & 3.0       & 900 & $1.0\times1.0\times1.0$ \\
             & Philips & MPRAGE          & 7       & 2.9--3.0  & 900 & $1.0\times1.0\times1.0$ \\
    OASIS--3 & Siemens & MPRAGE          & 2400    & 3.1--3.2  & 1000 & $1.0\times1.0\times1.0$ \\
    OASIS--4 & Siemens & MPRAGE          & 2300    & 2.95      & 900 & $1.0\times1.0\times1.0$ \\
    \bottomrule
  \end{tabularx}
\end{table}

\subsection{MRI acquisition}
All data consisted of 3T 3D sagittal T1-weighted (T1w) acquisitions obtained from the Alzheimer’s Disease Neuroimaging Initiative (ADNI) database (\url{adni.loni.usc.edu}) \citep{jack2008,jack2010b} and Open Access Series of Imaging Studies (OASIS) \citep{lamontagne2019,koenig2020} datasets. 
Data from ADNI--GO/2/3/4 and OASIS--3/4 were included. Accelerated images (factor=2) were collected for ADNI, and fully sampled images were collected for OASIS to maximize sample size and capture data from diverse imaging protocols.
Vendor-specific implementations included Siemens MPRAGE, GE IR-FSPGR and MPRAGE, and Philips MPRAGE (IR-TFE). Representative parameters (TR, TE, TI, voxel size) by dataset and vendor were summarized in Table~\ref{tab:mri_params}. The reported TR values in the table followed vendor-specific conventions and were not directly comparable across vendors. For Philips in ADNI--GO/2, TI was protocol-based because converted headers sometimes reported 0.

\begin{table}[t]
  \captionsetup{position=top}
  \caption{Participant demographics (n$=$1988) by scanner vendors.}
  \label{tab:demographics}
  \begin{threeparttable}
    \setlength{\tabcolsep}{6pt}
    \renewcommand{\arraystretch}{1.12}
    \begin{tabular*}{\linewidth}{@{\hspace{8pt}}@{\extracolsep{\fill}} l c c c c @{\hspace{8pt}}}
      \toprule
      Vendor & Diagnosis & n & Age (year) & Sex (F/M) \\
      \midrule
      Siemens & AD & 575  & 76.3 $\pm$ 7.6 & 274 / 301 \\
              & CN & 1123 & 69.8 $\pm$ 9.0 & 684 / 439 \\
      GE      & AD & 69   & 75.3 $\pm$ 8.0 & 25 / 44 \\
              & CN & 103  & 73.2 $\pm$ 7.3 & 58 / 45 \\
      Philips & AD & 52   & 73.7 $\pm$ 7.7 & 25 / 27 \\
              & CN & 66   & 73.3 $\pm$ 8.2 & 43 / 23 \\
      \bottomrule
    \end{tabular*}
    \vspace{2mm}
    \begin{tablenotes}[para,flushleft]
      \footnotesize
      Values are mean $\pm$ SD. All available participants were retained after preprocessing; no subjects were duplicated.
    \end{tablenotes}
  \end{threeparttable}
\end{table}

\subsection{Participants}
We analyzed $N=1{,}988$ T1w scans from the datasets described above. After preprocessing, the final sample included 696 AD patients and 1,292 CN individuals. Demographics and sample distributions by MR manufacturer (Siemens, GE, Philips) are summarized in Table~\ref{tab:demographics}. All participants were unique, with no duplicate subjects retained. Unless otherwise specified, primary analyses used the large-scale multi-cohort datasets, and manufacturer-specific analyses were conducted on subsets constructed by matching AD and CN counts across manufacturers to the smallest vendor (Philips).

\subsection{Cortical surface processing}
\begin{figure}[h]
\centering
\includegraphics[width=0.9\textwidth]{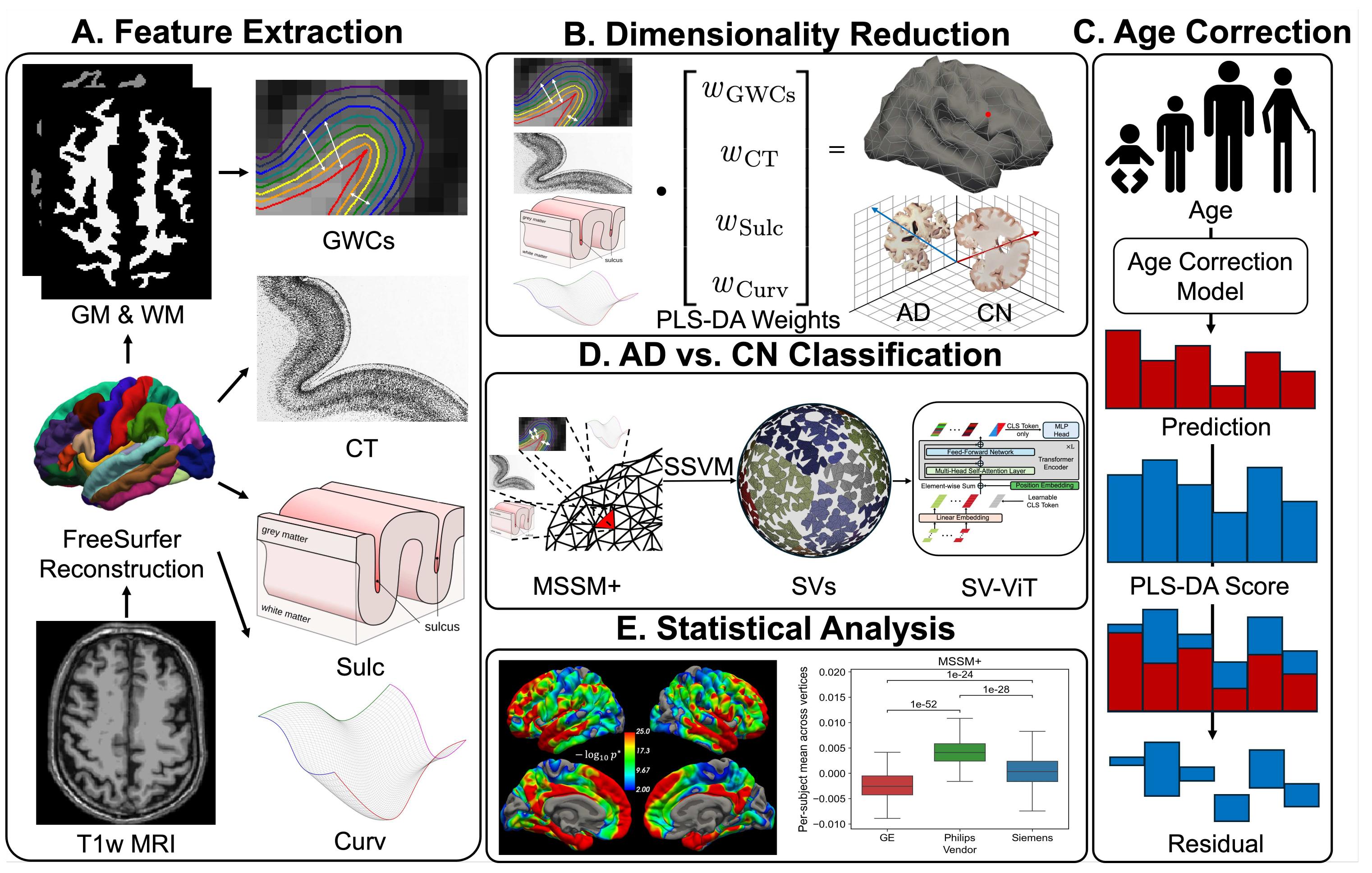}
\caption{\textbf{Overview of the MSSM+ procedure and classification using SSVM and SV-ViT.} GM, gray matter; WM, white matter; GWCs, gray-to-white matter contrasts; CT, cortical thickness; Sulc, sulcal depth; Curv, cortical curvature; PLS-DA, partial least squares-discriminant analysis; SVs, supervertices; AD, participants with Alzheimer's disease; CN, cognitively normal participants;}
\label{fig:pipeline}
\end{figure}

The overall MSSM+ analysis pipeline, including preprocessing, feature extraction, dimensionality reduction, age correction, group comparison, and classification, is summarized in Fig.~\ref{fig:pipeline}. Cortical surfaces were reconstructed using FreeSurfer v7.4.1 \citep{fischl2004}. All scans underwent standard FreeSurfer quality control; scans with significant reconstruction errors were excluded. Vertex-wise cortical features (GWCs, CT, Sulc, Jacob, Curv, and LGI) were extracted for each participant and mapped to the fsaverage surface. Following the original MSSM procedure \citep{jang2022}, GWCs were computed as multiple depth-wise gray–white intensity contrasts. Gray matter intensities were sampled at 20\%, 40\%, 60\%, and 80\% cortical depth, and white matter intensities were sampled 0.5\,mm and 1\,mm beneath the gray–white boundary. Ratios of these gray and white matter intensities yielded eight vertex-wise GWC maps per subject. This multiscale sampling quantified microstructural contrast at the cortical boundary, providing sensitivity to alterations associated with age and disease \citep{salat2009,salat2011,westlye2010}.

\subsection{Dimensionality reduction and age correction}
Because the features were multicollinear, PLS-DA \citep{lee2018} was applied at each vertex to derive a latent component that maximally covaried with the diagnosis label (AD vs. CN). For a single vertex-wise feature (e.g., CT alone), PLS-DA was not needed. To reduce age-related effects, vertex-wise residualization \citep{zhang2023} was performed using polynomial regression models. Age modeling was conducted using an independent subset of 150 CN participants similar to previous work \citep{adaszewski2013,tosun2016,amoroso2018,li2021}. Specifically, we sampled 25 CN participants from each vendor-by-sex stratum (3 vendors $\times$ 2 sexes = 150). The residuals were used in classifications and statistical analyses. Data from these participants were not reused in any subsequent statistical analyses or classification studies.
%In determining the sample size for age modeling, we referred to relevant literature and found that most literature uses less than 150 participants when preparing an independent subset for age correction modeling \citep{adaszewski2013,tosun2016,amoroso2018,li2021}. We therefore selected 150 as a pragmatic subset size. 

\subsection{Surface supervertex mapping}
\begin{figure}[H]
  \captionsetup{justification=raggedright, singlelinecheck=false}
  \centering
  \includegraphics[width=\linewidth]{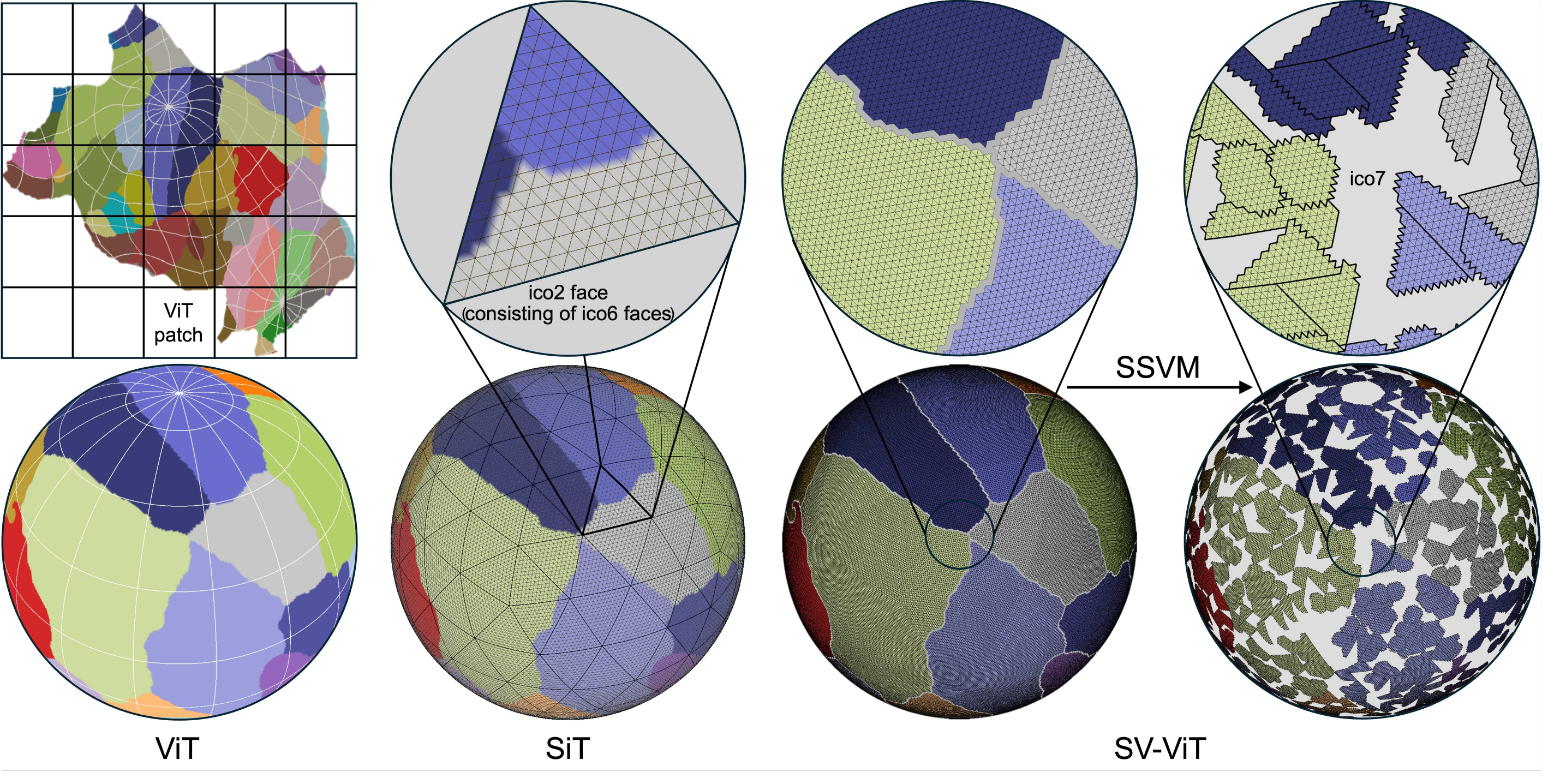}
  \caption{
  \textbf{Patch tokenization method (SSVM) for Supervertex Vision Transformer (SV-ViT).}
  }
  \label{fig:fig2}
\end{figure}

We developed SSVM (Fig.~\ref{fig:fig2}) to allow anatomically informed inter- and intra-ROI attention and alleviate excessive computational burden by defining finer-grained patches, or SVs. Vision Transformer (ViT) \citep{dosovitskiy2021} partitions 2D images into patches, which cannot directly represent cortical geometry. Surface Vision Transformer (SiT) \citep{dahan2022} adapts this concept using uniform icosahedral subdivision on the icosphere (ico6$\rightarrow$ico2) to produce equal-area patches that preserve geometry but ignore anatomical boundaries. In contrast, SV-ViT applied the proposed SSVM. The fsaverage icosphere (ico7) surface contains hundreds of thousands of triangular faces per hemisphere, making global self-attention computationally prohibitive. Faces belonging to the medial wall were excluded before SV construction. A face was treated as a boundary face when its three vertices had mixed ROI labels. The boundary faces were excluded before breadth-first search (BFS) initialization to prevent overlapping ROI assignment. We fixed the number of SVs to 1{,}280 per hemisphere (the number of ico3 faces), ensuring balanced, topology-preserving aggregation of cortical information. For each ROI, the number of patches was assigned in proportion to ROI size. Patch size, defined as the number of faces per patch, was chosen separately for each hemisphere as the largest feasible common value such that all patches in that hemisphere had the same number of faces.

\subsection{Supervertex Vision Transformer}
\begin{figure}[H]
  \captionsetup{justification=raggedright, singlelinecheck=false}
  \centering
  \includegraphics[width=\linewidth]{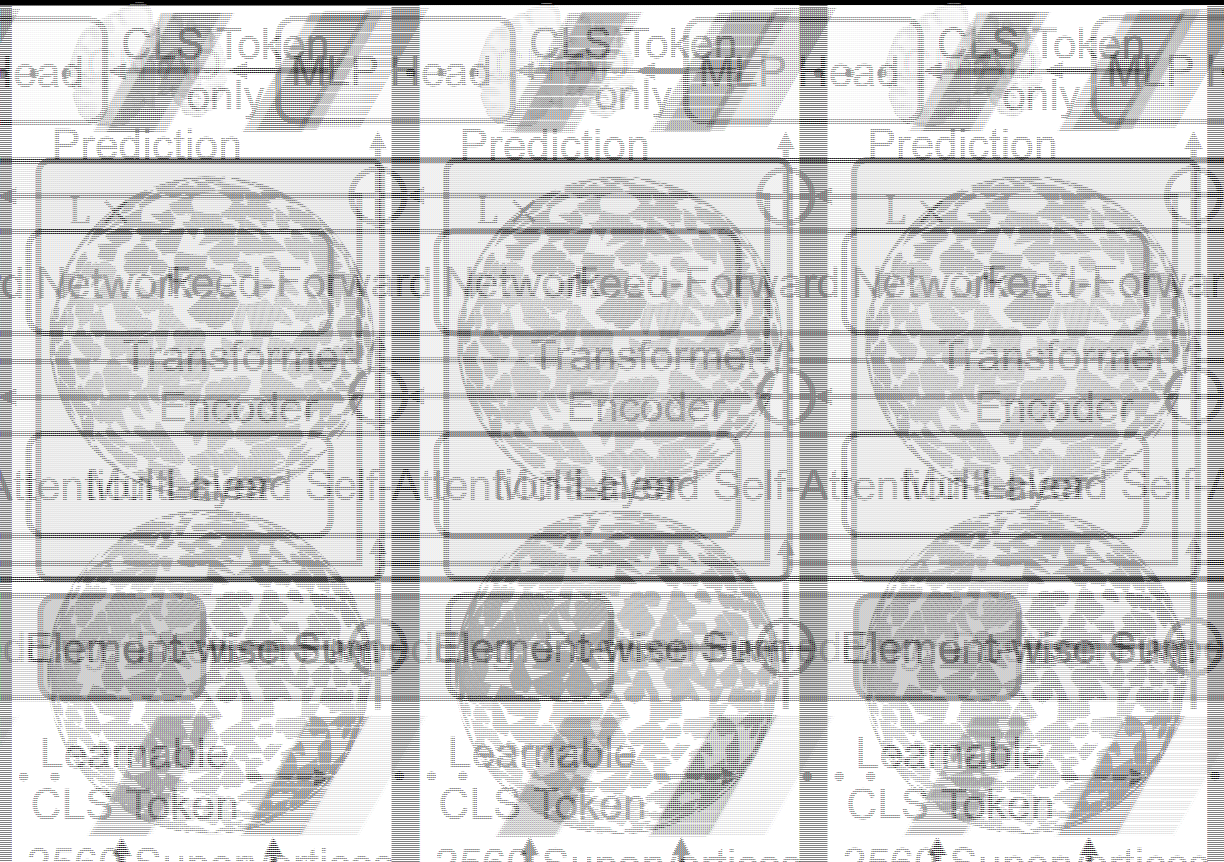}
  \caption{
  \textbf{Architecture of the proposed Supervertex Vision Transformer (SV-ViT).}}
  \label{fig:fig3}
\end{figure}

% 캡션들 짧게 하고 본문으로 옮기기 <- 제목만 남기고 본문으로 or 본문에도 적거나 - figure2,3 마찬가지
% We propose SV-ViT, which

% SV-ViT 설명, 및 내부 구조 설명

%Attention captures relational dependencies among supervertices, whereas position embedding encodes  %position embedding, supervertex 역할 추가, discussion에도 한번 더
% 어텐션차원, 포지션임베딩 차원 둘이 트레이드오프관계이다. <- discussion

We designed SV-ViT (Fig.~\ref{fig:fig3}), which extended SiT by applying the proposed SSVM (Fig.~\ref{fig:fig2}) to enable anatomically-informed learning. SSVM processed cortical surface maps into SVs, which were then used as input to SV-ViT, thereby enhancing efficiency and interpretability for surface-based learning. The SVs were linearly embedded, and a learnable class (CLS) token of the same dimension was appended to the token sequence. A position embedding was added to each token. Then the tokens passed through stacked Transformer encoder blocks, each comprising a multi-head self-attention layer and a feed-forward network. The attention mechanism captured relational dependencies among SVs, whereas position embedding preserved local cortical topography within SVs. From the output tokens, only the CLS token was fed to a multi-layer perceptron (MLP) head to generate a logit for AD classification.

\subsection{AD vs. CN classification}
All the classification analyses followed a standard protocol. For all models, the inputs were the age-corrected residuals. The tasks were formulated as binary classification (AD vs. CN) and evaluated using stratified 4-fold cross-validation, with subject-wise folds balanced by diagnosis. Within the training data, a further stratified split reserved 10\% of subjects for validation. When multiple models or feature configurations were compared in the same subject subset, we reused identical folds and partitions across methods. The models were optimized with the AdamW optimizer and a cosine annealing learning rate scheduler to minimize the binary cross-entropy loss. The classification performance was assessed on the held-out test folds. The evaluation metrics were the area under the receiver operating characteristic curve (AUROC) and the area under the precision--recall curve (AUPRC). We used AUROC as the primary metric and AUPRC as the secondary metric for selecting MSSM+.
We assessed all 63 combinations of vertex-wise cortical features (GWCs, CT, Sulc, Jacob, Curv, and LGI) by conducting AD detection with SV-ViT for each combination as input. We assessed the cross-vendor robustness of MSSM+ across MR manufacturers by comparing its classification performance with CT, GWCs, and MSSM. Siemens, GE, and Philips scans were evaluated.
Recent surface-based models, including HGNN+ \citep{feng2022}, SplineCNN \citep{fey2018}, SpiralNet++ \citep{gong2019}, and SiT \citep{dahan2022}, were applied to AD vs. CN classification with MSSM+ as input.
Ablation studies were performed within the proposed framework. Incremental feature integration was examined by performing SV-ViT for AD vs. CN classification while adding the constituent MSSM+ features to the input one at a time in order of their relative importance. Leave-one-out analyses were then carried out by conducting SV-ViT for AD vs. CN classification, removing a single component (either an MSSM+ feature or the SV representation) at a time and keeping all remaining components unchanged.

\subsection{Statistical analysis}
Group-level analyses were performed on age-corrected residuals, as described above, to estimate the effect of AD diagnosis on CT, GWCs, MSSM, and MSSM+. Spatial smoothing was applied on the fsaverage surface with a Gaussian kernel of 10~mm full width at half maximum (FWHM) only for group-level analyses. We used vertex-wise General Linear Models (GLMs) (FreeSurfer v7.4.1). Per-vertex \textit{p}-values were reported on pial surface maps (called significance maps). Permutation-based cluster-wise correction for multiple comparisons was performed with 10{,}000 permutations, using a cluster-forming threshold of 0.01 and a cluster-wise corrected threshold of 0.05. Only the vertices that survived after correction were colored on the significance maps. Primary and manufacturer-specific analyses were conducted.
Differences in subject-wise means of biomarkers (CT, GWCs, MSSM, and MSSM+) across MR manufacturers were investigated using Welch’s ANOVA, which accommodates unequal variances and unequal sample sizes. Accordingly, this analysis was carried out on the entire dataset rather than on vendor-specific subsets. When the omnibus Welch test was significant, pairwise Welch’s \textit{t}-tests between MR manufacturers (Siemens, GE, and Philips) were performed with Holm adjustment using a threshold of 0.05 for multiple comparisons.

\section{Results}

\subsection{Effect of extending MSSM with additional cortical features}

\begin{table}[htbp]
  \captionsetup{justification=raggedright,singlelinecheck=false}
  \centering
  \begin{threeparttable}
  \caption{\textbf{AD vs. CN classification performance for different cortical features in the combined multi-vendor dataset.} MSSM+ (GWCs+CT+Sulc+Curv) achieved the highest AUROC and AUPRC.}
  \label{tab:maintable}
  \begin{tabularx}{\textwidth}{p{6cm} >{\centering\arraybackslash}X >{\centering\arraybackslash}X}
    \toprule
    Features & AUROC & AUPRC \\
    \midrule
    CT~\citep{fischl2000,fischl2004} & 0.8475 $\pm$ 0.0228 & 0.7402 $\pm$ 0.0113 \\
    GWCs~\citep{salat2009} & 0.8660 $\pm$ 0.0187 & 0.7653 $\pm$ 0.0232 \\
    MSSM~\citep{jang2022} & 0.9166 $\pm$ 0.0103 & 0.8588 $\pm$ 0.0186 \\
    MSSM+ & \textbf{0.9312} $\pm$ \textbf{0.0051} & \textbf{0.8865} $\pm$ \textbf{0.0104} \\
    \bottomrule
  \end{tabularx}
  \begin{tablenotes}[para,flushleft]
    \footnotesize
    Values are reported as mean $\pm$ standard deviation across 4-fold cross-validation on held-out test folds.
  \end{tablenotes}
  \end{threeparttable}
\end{table}

\begin{figure}[H]
  \captionsetup{justification=raggedright,singlelinecheck=false}
  \centering
  \includegraphics[width=0.85\linewidth]{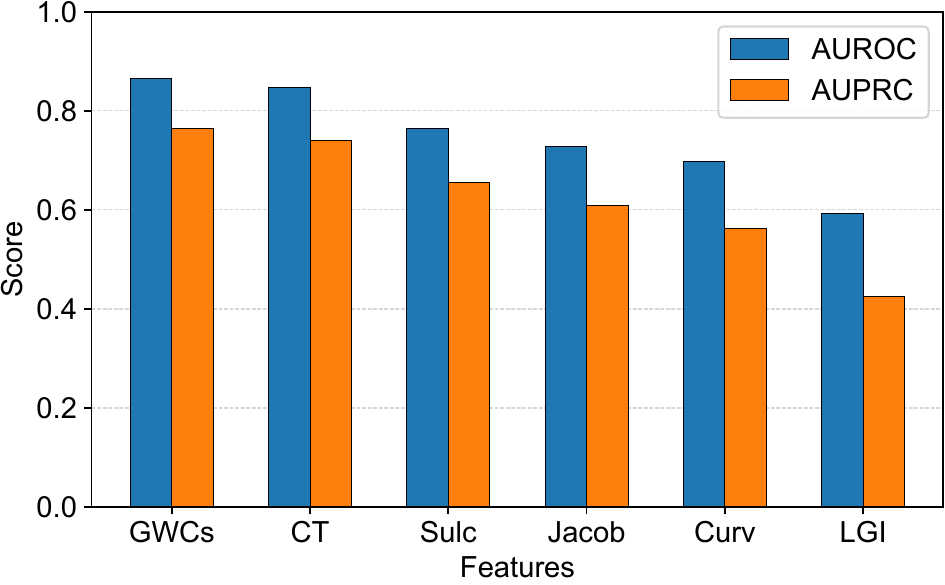}
  \caption{\textbf{AD vs. CN classification performance of individual cortical features using SV-ViT.} Metrics were averaged across a 4-fold cross-validation on held-out test folds. Bars show AUROC and AUPRC for each feature: GWCs, CT, Sulc, Jacob, Curv, and LGI.}
  \label{fig:fig4}
\end{figure}

We performed AD vs. CN classification using SV-ViT on large-scale multi-cohort datasets across all 63 combinations of cortical features (GWCs, CT, Sulc, Jacob, Curv, and LGI) to evaluate extensions of MSSM. Across all 63 feature combinations, MSSM+ achieved the highest AUROC and AUPRC (Supplementary Table~S1). As shown in Table~\ref{tab:maintable}, MSSM+ (GWCs+CT+Sulc+Curv) achieved a mean AUROC of 0.93 and an AUPRC of 0.89, representing improvements of 1\%p in AUROC and 3\%p in AUPRC over MSSM. Among individual features, GWCs showed the strongest baseline, followed by CT, Sulc, Jacob, Curv, and LGI, in terms of AUROC and AUPRC (Fig.~\ref{fig:fig4}).

\begin{figure}[htbp]
  \captionsetup{justification=raggedright,singlelinecheck=false}
  \centering
  \includegraphics[width=\linewidth]{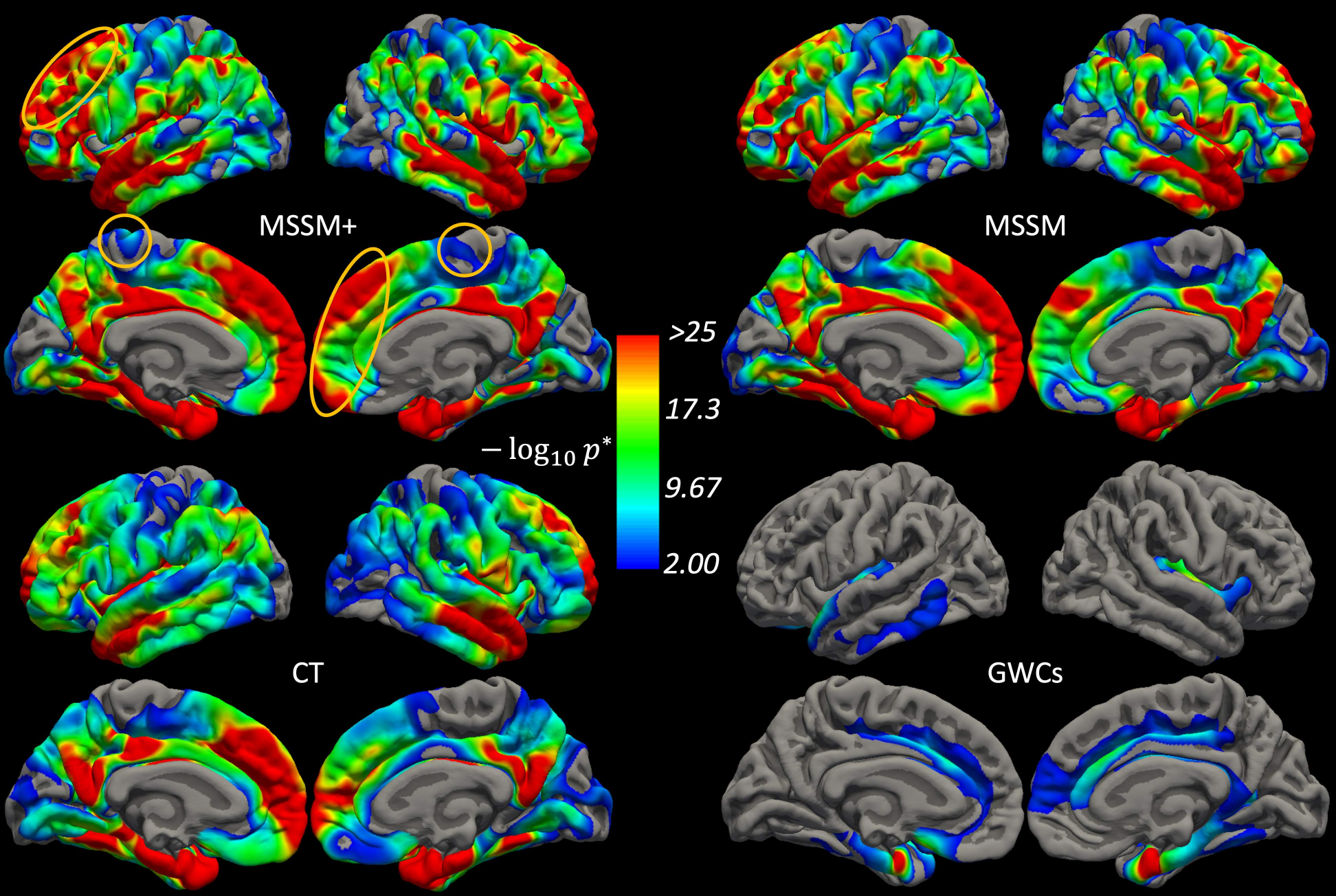}
  \caption{
  \textbf{Significance maps from group-level analysis of AD vs. CN in the combined multi-vendor dataset.}}
  \label{fig:fig5}
\end{figure}

Significance maps from the vertex-wise group-level analyses illustrated diagnosis-related differences across CT, GWCs, MSSM, and MSSM+ (Fig.~\ref{fig:fig5}). MSSM already showed broader and stronger effects than CT and GWCs. MSSM+ further extended these effects, with greater significance in the prefrontal and medial frontal cortices and more extensive significant regions in the paracentral cortex than in MSSM.

\subsection{Generalizability across MR manufacturers}
\begin{table}[htbp]
  \captionsetup{justification=justified}
  \centering
  \small
  \begin{threeparttable}
  \setlength{\tabcolsep}{4pt}
  \renewcommand\arraystretch{1.05}
  \newcommand{\numpm}[2]{\mbox{$#1 \,\pm\, #2$}}
  \caption{AD vs. CN classification performance by MR manufacturer and features. Vendor-specific subsets were down-sampled to match the smallest vendor (Philips) in sample size and AD:CN ratio.}
  \label{tab:generalization}
  \setlength{\tabcolsep}{2.75pt}
  \begin{tabularx}{\linewidth}{l c *{4}{>{\centering\arraybackslash}X} }
    \toprule
    & & \multicolumn{4}{c}{Features} \\
    \cmidrule(lr){3-6}
    Vendor & Metric & CT & GWCs & MSSM & MSSM+ \\
    \midrule
    Siemens & AUROC & \numpm{0.8433}{0.0256} & \numpm{0.8356}{0.0272} & \numpm{0.9100}{0.0137} & \numpm{0.9307}{0.0124} \\
            & AUPRC & \numpm{0.7163}{0.0429} & \numpm{0.7398}{0.0424} & \numpm{0.8465}{0.0259} & \numpm{0.8847}{0.0246} \\
    GE      & AUROC & \numpm{0.5664}{0.0997} & \numpm{0.8959}{0.0166} & \numpm{0.9172}{0.0169} & \numpm{0.9475}{0.0152} \\
            & AUPRC & \numpm{0.4920}{0.0824} & \numpm{0.8408}{0.0178} & \numpm{0.8743}{0.0363} & \numpm{0.9228}{0.0268} \\
    Philips & AUROC & \numpm{0.5892}{0.0831} & \numpm{0.9029}{0.0332} & \numpm{0.9142}{0.0247} & \numpm{0.9292}{0.0210} \\
            & AUPRC & \numpm{0.5838}{0.1117} & \numpm{0.8659}{0.0723} & \numpm{0.8745}{0.0653} & \numpm{0.8908}{0.0623} \\
    \bottomrule
  \end{tabularx}
  \begin{tablenotes}[para,flushleft]
    \footnotesize
    Values are reported as mean $\pm$ standard deviation across 4-fold cross-validation on held-out test folds.
  \end{tablenotes}
  \end{threeparttable}
\end{table}

\begin{figure}[H]
  \captionsetup{justification=raggedright}
  \centering
  \includegraphics[width=\linewidth]{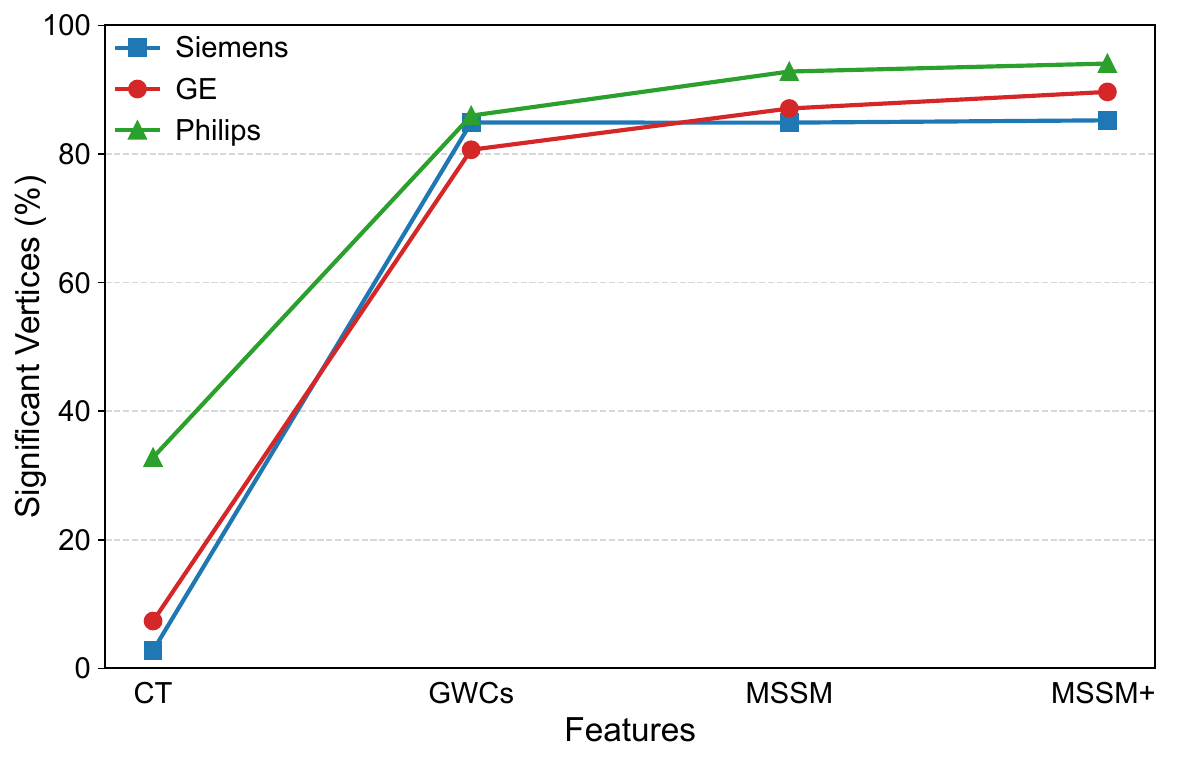}
  \caption{\textbf{Comparison of group-level AD vs. CN differences across MR manufacturers and features.}}
  \label{fig:fig6}
\end{figure}

In vendor-specific classifications, the AUROC and AUPRC of MSSM were higher and less variable across Siemens, GE, and Philips scanners than those of CT or GWCs. MSSM+ showed higher AUROC and AUPRC and lower variability across MR manufacturers than MSSM, achieving AUROC values of 0.93--0.94 across Siemens, GE, and Philips scanners (Table~\ref{tab:generalization}).

Vendor-specific group-level analyses were generally consistent with the classification results. For each manufacturer, the proportion of cortical vertices showing significant AD vs. CN differences increased from CT to GWCs to MSSM and MSSM+, indicating more extensive significant regions. Variability in the proportion of significant vertices across Siemens, GE, and Philips differed by features, as summarized in Fig.~\ref{fig:fig6}.

\begin{figure}[H]
  \captionsetup{justification=raggedright, singlelinecheck=false}
  \centering
  \includegraphics[width=\linewidth]{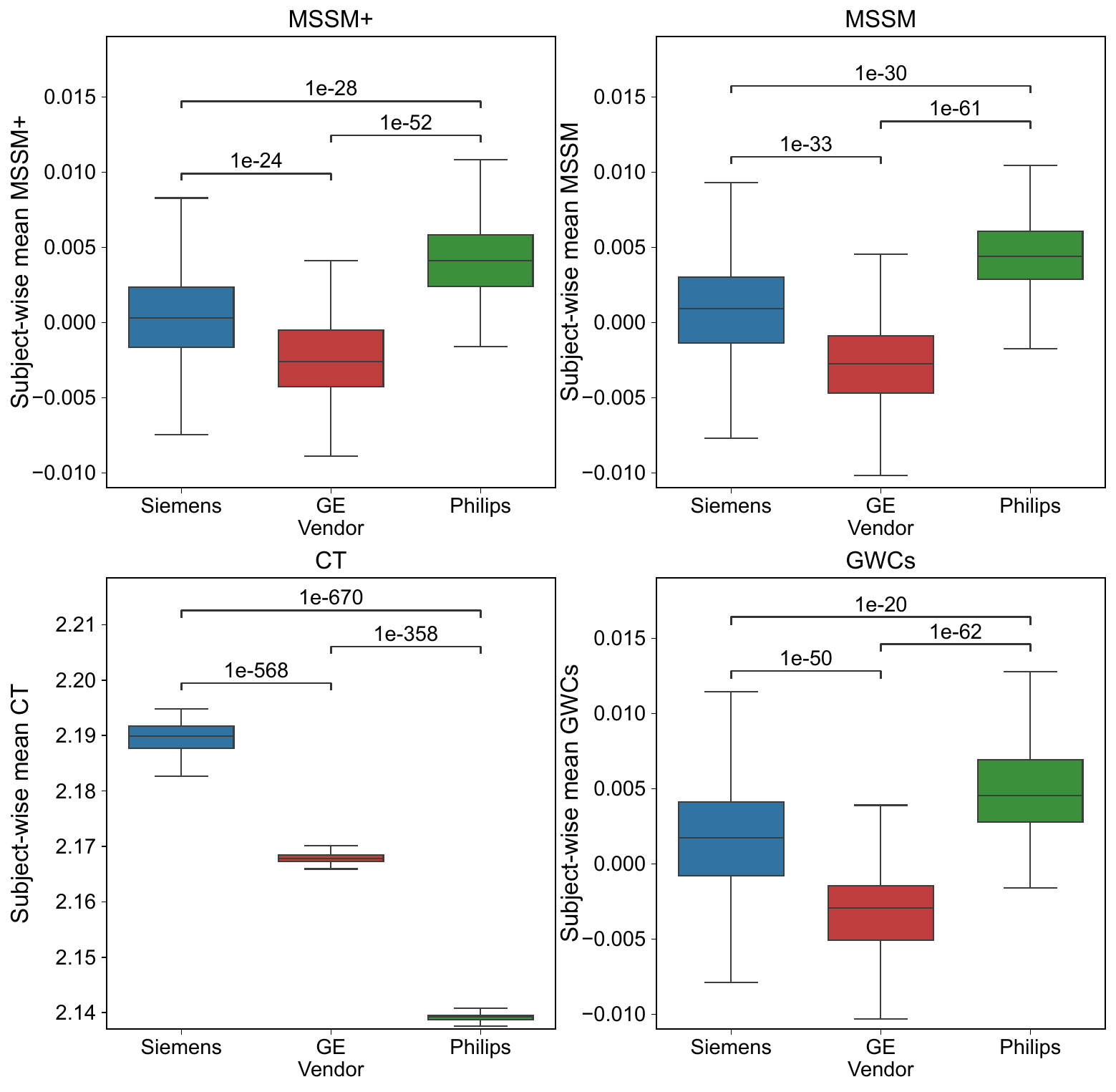}
  \caption{\textbf{Cross-vendor variability in subject-wise mean values of CT, GWCs, MSSM, and MSSM+.}}
  \label{fig:fig7}
\end{figure}

Analyses of subject-wise mean values using Welch’s ANOVA showed significant manufacturer effects for CT, GWCs, MSSM, and MSSM+ (all Holm-adjusted). Numbers above the brackets denote Holm-adjusted \textit{p}-values. Across measures, manufacturer-related variability decreased from CT to GWCs, then to MSSM, and then to MSSM+ (Fig.~\ref{fig:fig7}).

\subsection{Performance of SV-ViT with MSSM+ and ablation analyses}

\begin{table}[htbp]
  \captionsetup{justification=raggedright,singlelinecheck=false}
  \centering
  \begin{threeparttable}
  \caption{\textbf{AD vs. CN classification performance of recent surface-based models using MSSM+ as input.}}
  \label{tab:benchmarks}
  \begin{tabularx}{\textwidth}{@{}>{\raggedright\arraybackslash}X@{\hspace{6pt}}r@{\hspace{10pt}}r@{}}
    \toprule
    Models &
      \multicolumn{1}{@{\hspace{2pt}}c@{\hspace{12pt}}}{AUROC} &
      \multicolumn{1}{c}{AUPRC} \\
    \midrule
    Hypergraph Neural Network+ \citep{feng2022} & 0.8960 $\pm$ 0.0062 & 0.8133 $\pm$ 0.0128 \\
    Spline‑based Graph CNN \citep{fey2018} & 0.9015 $\pm$ 0.0366 & 0.8430 $\pm$ 0.0427 \\
    SpiralNet++ \citep{gong2019} & 0.9172 $\pm$ 0.0241 & 0.8575 $\pm$ 0.0412 \\
    Surface Vision Transformer \citep{dahan2022} & 0.9299 $\pm$ 0.0161 & 0.8796 $\pm$ 0.0289 \\
    Supervertex Vision Transformer & \textbf{0.9312} $\pm$ \textbf{0.0051} & \textbf{0.8865} $\pm$ \textbf{0.0104} \\
    \bottomrule
  \end{tabularx}
  \begin{tablenotes}[para,flushleft]
      \footnotesize
      Values are reported as mean $\pm$ standard deviation across 4-fold cross-validation on held-out test folds.
    \end{tablenotes}
  \end{threeparttable}
\end{table}

% reviewr1-1
In each hemisphere, approximately 2.9\% of the 327,680 faces were classified as boundary faces and excluded before BFS initialization. Excluding the medial wall, no vertices were excluded at this stage, because every vertex of an excluded boundary face was also shared by at least one retained neighboring face.

% 필요할 경우 - 약점 - 제외된 vertex 수
% To preserve ROI boundaries while enforcing an equal number of faces per patch, some vertices were not included in the final supervertex representation; approximately 24.8\% of vertices were left unused in each hemisphere.

% benchmark results
Using MSSM+ as the standard input, we compared SV-ViT with recent surface-based models, including HGNN+, SplineCNN, SpiralNet++, and SiT. SV-ViT and SiT, both Vision Transformer–based architectures, yielded the largest AUROC and AUPRC, with SpiralNet++ and SplineCNN showing intermediate performance, and HGNN+ having lower mean values but the second-smallest variability across folds. Among all models, SV-ViT showed the highest mean AUROC and AUPRC and the smallest standard deviations (Table~\ref{tab:benchmarks}).

\begin{table}[htbp]
  \captionsetup{justification=justified, singlelinecheck=false}
  \centering
  \setlength{\tabcolsep}{8pt}
  \caption{\textbf{Incremental ablation of cortical feature sets for AD vs. CN classification using SV-ViT.}}
  \label{tab:incremental}
  \begin{threeparttable}
  \begin{tabularx}{\textwidth}{l *{2}{>{\centering\arraybackslash}X}}
    \toprule
    Features                  & AUROC               & AUPRC               \\
    \midrule
    GWCs                      & 0.8660 $\pm$ 0.0187 & 0.7653 $\pm$ 0.0232 \\
    GWCs+CT                   & 0.9166 $\pm$ 0.0103 & 0.8588 $\pm$ 0.0186 \\
    GWCs+Sulc+CT              & 0.9230 $\pm$ 0.0066 & 0.8713 $\pm$ 0.0102 \\
    GWCs+Sulc+CT+Curv (MSSM+) & 0.9312 $\pm$ 0.0051 & 0.8865 $\pm$ 0.0104 \\
    \bottomrule
   \end{tabularx}
  \begin{tablenotes}[para,flushleft]
    \footnotesize
    Values are reported as mean $\pm$ standard deviation across 4-fold cross-validation on held-out test folds.
  \end{tablenotes}
  \end{threeparttable}
\end{table}

\begin{table}[htbp]
  \captionsetup{justification=justified, singlelinecheck=false}
  \centering
  \setlength{\tabcolsep}{8pt}
  \caption{\textbf{Leave-one-out ablation of MSSM+ with SV-ViT for AD vs. CN classification.} Each row reports performance when one component of the proposed framework (e.g., GWCs, CT, Sulc, Curv, or the SV representation) was removed. $\Delta$AUROC and $\Delta$AUPRC denote the absolute decrease in performance relative to the full framework.}
  \label{tab:leaveoneout}
  \begin{threeparttable}
  \begin{tabular}{lcccc}
    \toprule
    Components & AUROC & $\Delta$AUROC & AUPRC & $\Delta$AUPRC \\
    \midrule
    MSSM+    & 0.9312 $\pm$ 0.0051 & - & 0.8865 $\pm$ 0.0104 & - \\
    w/o Curv & 0.9230 $\pm$ 0.0066 & 0.0082 & 0.8713 $\pm$ 0.0102 & 0.0153 \\
    w/o Sulc & 0.9236 $\pm$ 0.0100 & 0.0076 & 0.8674 $\pm$ 0.0198 & 0.0192 \\
    w/o CT   & 0.9208 $\pm$ 0.0064 & 0.0105 & 0.8612 $\pm$ 0.0077 & 0.0253 \\
    w/o GWCs & 0.8769 $\pm$ 0.0121 & 0.0543 & 0.7983 $\pm$ 0.0080 & 0.0882 \\
    w/o SV   & 0.8825 $\pm$ 0.0273 & 0.0487 & 0.8049 $\pm$ 0.0580 & 0.0816 \\
    \bottomrule
  \end{tabular}
  \begin{tablenotes}[para,flushleft]
    \footnotesize
    Values are reported as mean $\pm$ standard deviation across 4-fold cross-validation on held-out test folds.
  \end{tablenotes}
  \end{threeparttable}
\end{table}

Feature-wise AD detection using SV-ViT showed that GWCs had the highest AUROC and AUPRC among individual cortical features, followed by CT, Sulc, Jacob, Curv, and LGI (Fig.~\ref{fig:fig4}).

Incremental ablation for feature integration showed monotonic gains in AUROC and AUPRC as additional features were added to GWCs (Table~\ref{tab:incremental}). Adding CT improved the metrics, yielding MSSM (GWCs+CT), and subsequent inclusion of Sulc and Curv yielded smaller but consistent increases.

Leave-one-out ablations of MSSM+ further quantified the contribution of each component (Table~\ref{tab:leaveoneout}). Removing GWCs yielded the largest reductions in metrics, followed by removing the SV representation, then CT, whereas removing Sulc or Curv produced smaller performance decreases.

\section{Discussion}
In this study, using large-scale multi-cohort datasets, we proposed MSSM+ and demonstrated that MSSM (GWCs+CT) could be improved by incorporating the additional cortical features Sulc and Curv. In addition, MSSM+ showed generalizability across MR manufacturers. We also designed SSVM and SV-ViT and showed that anatomically informed learning on SVs could improve AD vs. CN classification performance.

MSSM+ (GWCs+CT+Sulc+Curv) yielded higher classification performance than MSSM, with approximately 3\%p higher AUPRC and 1\%p higher AUROC, as well as lower standard deviations across folds. In the group-level analyses, MSSM+ also showed greater spatial extent and stronger statistical significance than MSSM. Orange ellipses highlight regions where MSSM+ showed significant effects in the superior, middle, and medial frontal cortices and more extensive regions in the paracentral cortices compared with MSSM (Fig.~\ref{fig:fig5}). These findings suggest that extending the MSSM with Sulc and Curv yields measurable gains beyond CT and GWCs alone. These findings are also consistent with recent MSSM work suggesting that additional cortical surface features may further improve performance \citep{jang2024}. Although Sulc and Curv showed smaller individual contributions than GWCs, they may provide complementary geometric information beyond tissue contrast and cortical thickness. The dominant contribution of GWCs is likewise consistent with prior evidence that GWC alterations are closely associated with neurodegeneration and cognitive decline across the AD continuum \citep{xu2024}. In particular, these features characterize sulcal and local surface-shape variation that may reflect cortical degeneration not fully captured by GWCs or CT alone \citep{schaer2008,thompson2004,hogstrom2013}. 
Jacob showed higher single-feature classification performance than Curv, and, except when added to the full MSSM+ feature combination, its inclusion generally increased performance. This pattern suggests that excessive feature complexity can degrade performance. Consistent with this, Jacob quantifies local tissue expansion and contraction and has been widely used in deformation-based morphometry to detect subtle shape changes not captured by volumetric measures alone \citep{gaser2001,ashburner2000}.
LGI reduced classification performance when combined with other features, which we interpreted as indicating that LGI was not well suited to capture neurodegeneration; the fact that LGI has been reported to be primarily influenced by neurodevelopmental factors supports this interpretation \citep{schaer2008}.
Across vendor-specific subsets, MSSM+ and MSSM showed similar AD vs. CN classification performance. In group-level analyses, both MSSM+ and MSSM identified more extensive regions of significant AD vs. CN differences than CT and GWCs. Across vendors, variability in subject-wise mean MSSM+ values, followed by MSSM, was lower than that of CT and GWCs, indicating that MSSM+ and MSSM were more consistent across MR manufacturers.
SSVM enables inter- and intra-ROI-informed learning and alleviates computational complexity by defining finer-grained patches, or SVs. The attention mechanism captures inter-ROI dependencies among SVs. In contrast, the position embedding retains the intra-ROI information within SVs. Increasing the attention dimension therefore increases computational cost while also increasing the capacity to capture inter-regional dependencies. In our design, the attention dimension and position embedding dimension are constrained such that increasing one necessarily requires reducing the other. By controlling the number and size of SVs, our design optimizes an effective balance between computational burden and the richness of inter-regional relations.
SV-ViT and SiT, ViT-based models, exhibited higher classification performance than SpiralNet++, SplineCNN, and HGNN+, indicating that attention between cortical regions was pivotal for AD classification. The uniform icosahedral patches of SiT serve to partition the cortical surface into equally sized inputs for the model. In contrast, the SVs in SV-ViT, in addition to serving this role, enable ROI-informed learning. The SVs contributes to the performance improvement of SV-ViT  (Table~\ref{tab:leaveoneout}).

\subsection{Limitations}
For SSVM, we adopted a rule‑based, ROI‑constrained BFS to form equal‑face, edge‑connected, non‑overlapping supervertices within a single anatomical ROI. Only faces whose three vertices shared an identical anatomical ROI were included; all others were excluded to prevent ROI mixing. This yields a small, intentional under‑coverage concentrated at ROI borders and reflects design constraints rather than model failure. The vendor distribution was imbalanced (Siemens was the largest), raising the possibility that performance gains could be driven primarily by Siemens. Accordingly, we conducted vendor-specific analyses, including classification (Table~\ref{tab:generalization}), group-level analyses (Fig.~\ref{fig:fig6}), and cross-vendor variability analyses (Fig.~\ref{fig:fig7}). These analyses showed that the advantages of MSSM+ were observed not only in the full sample (Table~\ref{tab:maintable}) but also consistently across all vendors. We also used PLS-DA for vertex-wise dimensionality reduction; while computationally efficient here, multi‑component or supervised variants may capture additional distributed effects. Age differences between groups are an inherent concern in AD studies. We mitigated this by fitting vertex‑wise polynomial age adjustment models on a small, balanced subset and applying the learned coefficients to the entire dataset; residual confounding may remain.

\subsection{Future directions}
Future directions include replacing the heuristic with a learned, connectivity-preserving assignment that increases coverage while maintaining ROI fidelity and size balance in SSVM. This can be implemented on the cortical surface mesh adjacency graph using learned graph clustering instead of the current rule-based BFS. A vertex-based approach will prevent boundary faces consisting of vertices with mixed ROI labels and may improve ROI fidelity. We plan to refine architectures that more explicitly leverage cortical surface topology and vertex spatial relationships. Beyond the AD-CN classification, this framework can be evaluated in mild cognitive impairment (MCI) to examine progression to AD, considering that prior work integrating structural features maps ~\citep{jang2022} demonstrated potential for longitudinal application. Further subgroup analyses by disease severity will also be important for better understanding stage-specific patterns of disease-related change. An additional priority is to integrate amyloid-$\beta$ (A$\beta$) and tau biomarkers from CSF or PET, thereby testing alignment with the A/T/N framework and evaluating prediction of A$\beta$-positive and tau-positive status. We will further examine more refined age-modeling strategies and conduct sensitivity analyses to assess the robustness of the age-correction procedure. It will also be important to determine whether specific training-sample compositions, including vendor-balanced subsets, can further improve cross-vendor performance. Scanner harmonization across MR manufacturers will be examined in future studies. Finally, extending the approach to multimodal structural MRI, such as T2-weighted, fluid-attenuated inversion recovery, and diffusion imaging, may further improve generalizability and sensitivity to early microstructural change.

\section{Conclusion}
We proposed MSSM+, an extension of MSSM that integrates additional morphometric features at the vertex level. We also designed SSVM and SV-ViT, which use SVs from SSVM to enable anatomically informed learning. Together, these approaches improved performance in AD vs. CN classification for each MR vendor and yielded stronger, broader group contrast across large-scale multi-cohort datasets. These findings suggest that our approach may serve as a vendor-generalizable and accessible MRI biomarker for the detection of AD prior to CSF/PET confirmation.

\section*{CRediT authorship contribution statement}
{\noindent}Geonwoo Baek: Conceptualization, Methodology, Software, Formal analysis, Writing – original draft.\\
David H. Salat: Supervision, Validation, Writing – review \& editing.\\
Ikbeom Jang: Conceptualization, Supervision, Project administration, Writing – review \& editing.

\section*{Declaration of competing interest}
The authors declare that they have no known competing financial interests or personal relationships that could have influenced the work reported in this paper.

\section*{Funding}
\sloppy
This work was supported by the National Institutes of Health (NIH) grant (R21AG072431) of the U.S.; the National Research Foundation of Korea (NRF) grant funded by the Ministry of Science and ICT (MSIT) (RS-2024-00455720); the Korea National Institute of Health (KNIH) research projects (2024ER040700 \& 2025ER040300); the National Supercomputing Center with supercomputing resources including technical support (KSC-2024-CRE-0021 \& KSC-2025-CRE-0065); and Hankuk University of Foreign Studies Research Fund of 2025.

\section*{Ethics approval and consent to participate}
Data used in the preparation of this article were obtained from the ADNI \citep{jack2008,jack2010b} and OASIS databases \citep{lamontagne2019,koenig2020}. Both ADNI and OASIS obtained informed consent from all participants, and their study protocols were approved by the Institutional Review Boards of the participating institutions.

\begin{comment} # 원문
*Data used in preparation of this article were obtained from the Alzheimer’s Disease Neuroimaging Initiative
(ADNI) database (adni.loni.usc.edu). As such, the investigators within the ADNI contributed to the design
and implementation of ADNI and/or provided data but did not participate in analysis or writing of this report.
A complete listing of ADNI investigators can be found at:
http://adni.loni.usc.edu/wp-content/uploads/how_to_apply/ADNI_Acknowledgement_List.pdf
\end{comment}

\section*{Data availability}
The ADNI and OASIS data are available at \url{https://adni.loni.usc.edu} and \url{https://www.oasis-brains.org}, respectively. As such, the investigators within the ADNI contributed to the design
and implementation of ADNI and/or provided data but did not participate in analysis or writing of this report.
A complete listing of ADNI investigators can be found at:
\url{http://adni.loni.usc.edu/wp-content/uploads/how_to_apply/ADNI_Acknowledgement_List.pdf}

\section*{Code availability}
The code used in this study will be made publicly available upon publication at \url{https://github.com/labhai/MSSMplus}.

\section*{Acknowledgments}
Data collection and sharing for this project was funded by the Alzheimer's Disease Neuroimaging Initiative (ADNI) (National Institutes of Health Grant U01 AG024904) and DOD ADNI (Department of Defense award number W81XWH-12-2-0012). ADNI is funded by the National Institute on Aging, the National Institute of Biomedical Imaging and Bioengineering, and through generous contributions from the following: AbbVie, Alzheimer’s Association; Alzheimer’s Drug Discovery Foundation; Araclon Biotech; BioClinica, Inc.; Biogen; Bristol-Myers Squibb Company; CereSpir, Inc.; Cogstate; Eisai Inc.; Elan Pharmaceuticals, Inc.; Eli Lilly and Company; EuroImmun; F. Hoffmann-La Roche Ltd and its affiliated company Genentech, Inc.; Fujirebio; GE Healthcare; IXICO Ltd.; Janssen Alzheimer Immunotherapy Research \& Development, LLC.; Johnson \& Johnson Pharmaceutical Research \& Development LLC.; Lumosity; Lundbeck; Merck \& Co., Inc.; Meso Scale Diagnostics, LLC.; NeuroRx Research; Neurotrack Technologies; Novartis Pharmaceuticals Corporation; Pfizer Inc.; Piramal Imaging; Servier; Takeda Pharmaceutical Company; and Transition Therapeutics. The Canadian Institutes of Health Research is providing funds to support ADNI clinical sites in Canada. Private sector contributions are facilitated by the Foundation for the National Institutes of Health (www.fnih.org). The grantee organization is the Northern California Institute for Research and Education, and the study is coordinated by the Alzheimer’s Therapeutic Research Institute at the University of Southern California. ADNI data are disseminated by the Laboratory for Neuro Imaging at the University of Southern California. OASIS data were provided by the OASIS project \citep{lamontagne2019,koenig2020}, Principal Investigators: D. Marcus, R. Buckner, J. Csernansky, and J. Morris.

\section*{Declaration of generative AI and AI-assisted technologies in the writing process}
During the preparation of this work, the authors used ChatGPT (OpenAI) to improve readability and language clarity. The authors reviewed and edited all content, and take full responsibility for the content of the published article.

\bibliographystyle{elsarticle-harv}
\bibliography{references}

\clearpage
\includepdf[pages=-]{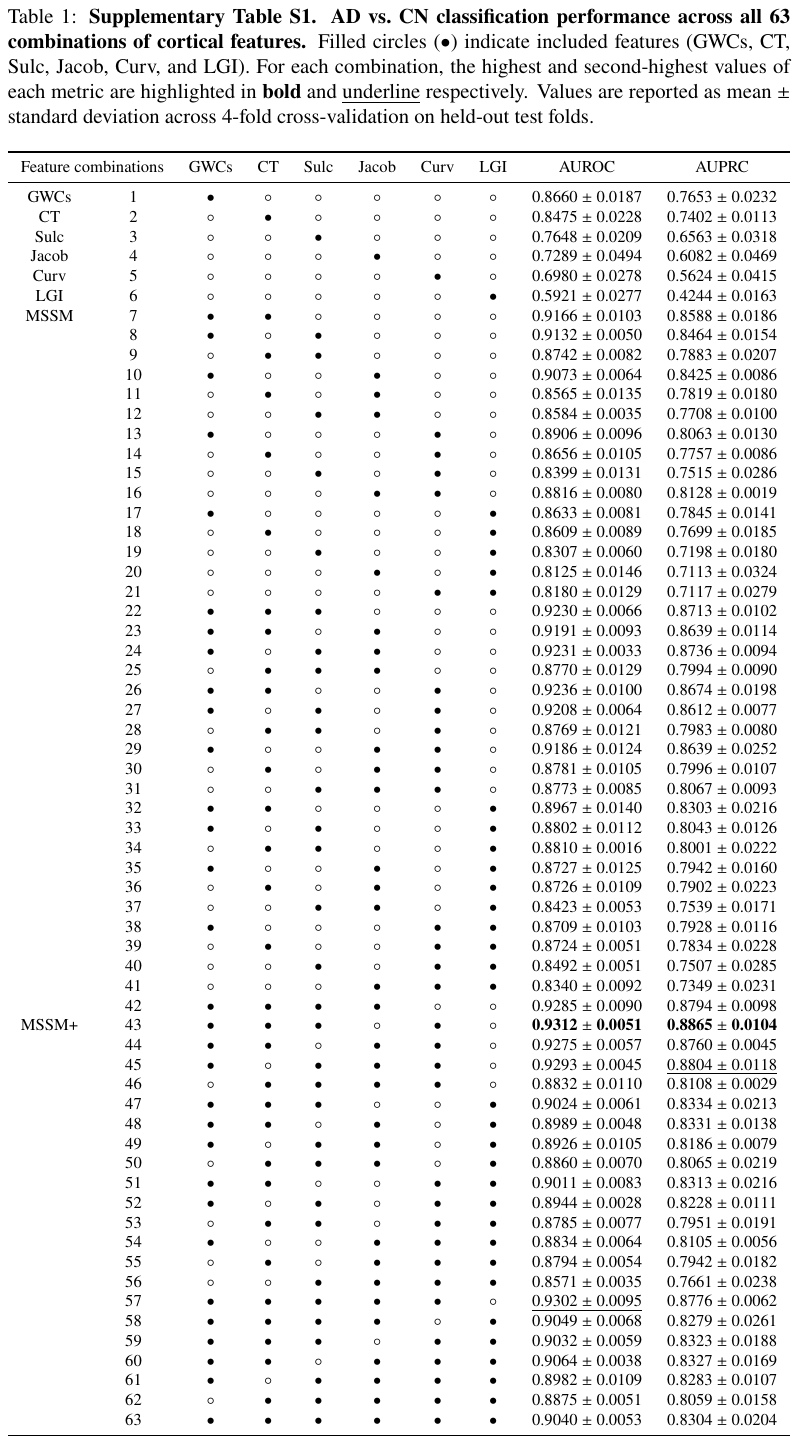}

\end{document}